# An Embarrassingly Simple Speed-Up of Belief Propagation with Robust Potentials

James M. Coughlan and Huiying Shen
The Smith-Kettlewell Eye Research Institute
2318 Fillmore St.
San Francisco, CA 94115
coughlan@ski.org , hshen@ski.org

## Abstract

We present an exact method of greatly speeding up belief propagation (BP) for a wide variety of potential functions in pairwise MRFs and other graphical models. Specifically, our technique applies whenever the pairwise potentials have been *truncated* to a constant value for most pairs of states, as is commonly done in MRF models with robust potentials (such as stereo) that impose an upper bound on the penalty assigned to discontinuities; for each of the $M$ possible states in one node, only a smaller number $m$ of compatible states in a neighboring node are assigned milder penalties. The computational complexity of our method is $O(mM)$, compared with $O(M^2)$ for standard BP, and we emphasize that the method is *exact*, in contrast with related techniques such as pruning; moreover, the method is very simple and easy to implement. Unlike some previous work on speeding up BP, our method applies both to sum-product and max-product BP, which makes it useful in any applications where marginal probabilities are required, such as maximum likelihood estimation. We demonstrate the technique on a stereo MRF example, confirming that the technique speeds up BP without altering the solution.

## 1. Introduction

Belief propagation (BP) has been applied in a wide variety of problems in computer vision and pattern recognition that use graphical models, including stereo [1], optical flow [2], tracking [3], shape matching [4,5], super-resolution [6], turbocodes [7], unwrapping 2D phase images [8], speech recognition [9] and protein folding [10]. While BP provides a tractable method for performing inference with a variety of graphical models, the computational complexity still scales with the square of the state space size (i.e. the joint state space of neighboring variables coupled by a pairwise potential), which prohibits many practical applications.

This paper introduces a novel and extremely simple technique that greatly speeds up BP without introducing any approximations. In this paper we will restrict the discussion to pairwise MRFs and pairwise BP, although the principle generalizes to factor graphs and factor BP [11] (as we describe in Sec. 7). Our technique applies only to graphical models whose variables have discrete (quantized) state spaces.

To introduce our technique, we first define the joint probability specified by a pairwise MRF:

$$P(x_1, \ldots, x_N) = \frac{1}{Z} \prod_{<ij>} f(x_i, x_j) \prod_i g(x_i) \tag{1}$$

where the product $\prod_{<ij>}$ is over all pairs of neighboring variables. Here we consider the case of a homogeneous MRF for simplicity, but we can easily generalize all the results in this paper to inhomogeneous MRFs, in which the potential functions depend on the nodes as well as the values of the nodes, i.e. $f_{ij}(x_i, x_j)$ and $g_i(x_i)$ instead of $f(x_i, x_j)$ and $g(x_i)$.

The speed-up applies whenever the pairwise potentials are robust, and specifically sparse in the sense that the potential value has been truncated over most of the state space: in other words, $f(x_i, x_j) = \bar{f}$ for most state pairs $(x_i, x_j)$, where $\bar{f}$ is a constant. (Again, this constant can depend on $i$ and $j$ if we have $f_{ij}(x_i, x_j)$ instead of $f(x_i, x_j)$.) We refer to this as a *sparse compatibility* property of the robust pairwise potential, meaning that ``incompatible'' pairs of states $x_i$ and $x_j$ have the same potential value $f(x_i, x_j) = \bar{f}$; see Fig. 1 for an illustration.

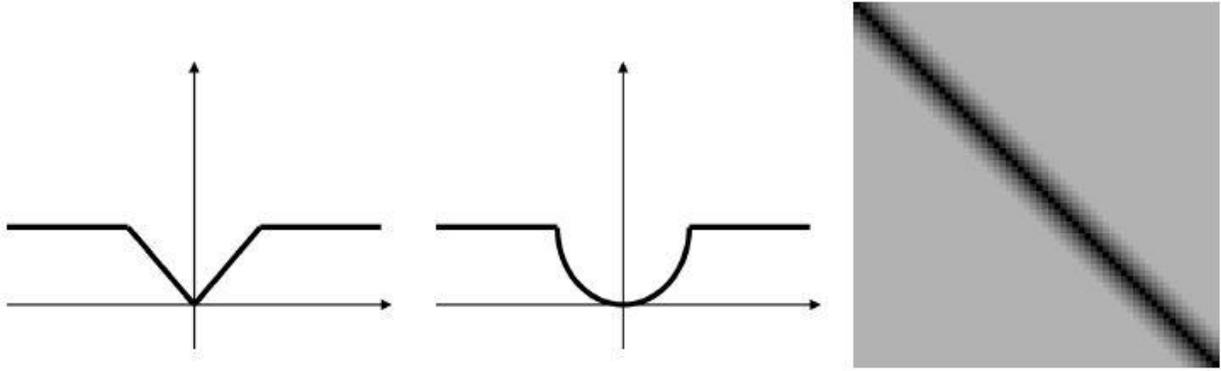

**Figure 1.** Illustration of sparse compatibility property of robust potentials. (a) and (b) show typical robust energy functions $E(x)$, where the corresponding pairwise potential function is given by $f(x_i, x_j) = e^{-E(x_i - x_j)}$. Note that the energy function is truncated at a maximum penalty value, and the energy function assumes this maximum value for most points in its domain. (c) shows $f(x_i, x_j)$ as a grayscale matrix ($x_i$ and $x_j$ are now discrete variables labeling the matrix row and column, and darker shading indicates higher potential values) showing that the potential has the same value, $\bar{f}$, for all incompatible state pairs (which are the majority of *all* state pairs).

Typically $\bar{f}$ is a small – but non-zero – number reflecting the fact that the MRF discourages (but does not prohibit) such state pairs. As an example, in MRF stereo we may set the pairwise smoothness potential to a small value whenever two neighboring disparities are sufficiently different: for instance, $f(x_i, x_j) = e^{-\alpha E(x_i - x_j)}$, where $E(x_i - x_j) = \min(|x_i - x_j|, T)$, which implies that $f(x_i, x_j) = \bar{f} = e^{-\alpha T}$ whenever $|x_i - x_j| > T$. (However, $\bar{f}$ can take any value and need not be small for our trick to apply to sum-product BP.)

## 2. Related Work

Several techniques have been advanced to increase the speed of BP. Perhaps the simplest speed-up technique is to ``prune'' the pairwise potential $f(x_i, x_j)$ by approximating its smaller entries as zeros, which allows one to streamline BP by ignoring most state pairs of neighboring variables in the message update calculations. While sometimes a convenient approximation, this pruning technique can cause problems if the state pairs being ignored actually have a non-negligible pairwise potential, in which case the solution determined by BP may be distorted by the approximation. (See further discussion of this approximation in Sec.'s 4 and 6.) Related pruning/beam search techniques [12,13,14] dynamically remove available states for each variable in the graphical model when its belief (i.e. current probability estimated by BP) is sufficiently low, but suffer from a similar problem: an important state may be removed prematurely in the course of running BP and lead to an incorrect solution.

A few exact BP speed-up techniques have been proposed. For max-product BP, the distance transform is one such technique [15,2] for MRFs with certain types of pairwise potential energies that are commonly used in computer vision and pattern recognition, such as an absolute value (linear) or quadratic energy function of the difference of neighboring pairs of variables, or a truncated version of these functions.

While the distance transform confers a massive speed-up for those models to which it is applicable, it has an important limitation: it applies to max-product BP but not sum-product BP. In some common problems such as maximum likelihood estimation or CRF parameter estimation [16,17], which require expectations to be estimated, sum-product BP is needed to compute belief marginals. (Indeed, for many applications in which marginal probabilities are not required, graph cuts often provide superior results to BP [18], and can be computed faster than BP.)

Two other speed-up methods are described in [2], multi-scale message updating and a bipartite graph (``checkerboard'') message update schedule. Both methods apply to MRFs formulated on a lattice (grid), and both are exact in the sense that they don't alter the fixed points of BP (although they may affect the precise results obtained, depending on if and how the message updates converge). Our technique can be applied in tandem with these methods, which is a practical way of speeding up sum-product BP for lattice MRF models such as stereo. (See Sec. 6.)

Other methods are available for speeding up sum-product BP, including the FFT and the box sum method, but both methods have important limitations. The FFT applies [11,7,2] to any model in which the pairwise potentials are functions of the difference of the two variables connected by the potential, and while it reduces the computational complexity of BP message updates from $O(M^2)$ to $O(M \log M)$ (where $M$ is the size of the state space), it may be too slow in practical applications. The box sum method [2] is a method that approximates a Gaussian potential function (or sum of Gaussians) by repeatedly convolving a small number of box filters, which is much less general than the FFT approach, and only approximate.

Our work is most related to [19], which describes a technique that both speeds up generalized BP and reduces the required memory storage. Like our technique, theirs exploits the sparse compatibility property of the MRF potentials; however, unlike ours, a speed-up is not guaranteed in general[1], but only for the case $\bar{f} = 0$ (i.e. the pairwise potential *forbids* incompatible states rather than merely penalizing them).

---

[1] This limitation is made explicit in the corrected version of the paper, available on the second author's website [19].

Finally, recent work [20] describes an exact speed-up technique for BP in which the potentials (typically higher-order than pairwise, and usually expressed in terms of factor graphs [11]) are linear combinations of their arguments (or non-linear functions of such linear combinations), based on changing variables in the message update equations. However, this method is typically not useful for pairwise MRFs, where speed-ups are still needed. Moreover, for some factor graphs this method will be inapplicable but our technique will still apply (see Sec. 7).

## 3. Problem Formulation

In standard sum-product (pairwise) BP [21], the message from node $i$ to node $j$ is denoted by $m_{ij}(x_j)$ and is updated according to the following equation:

$$m_{ij}^{new}(x_j) = \sum_{x_i} f(x_i, x_j) g(x_i) \prod_{k \in N(i) \backslash j} m_{ki}(x_i) \qquad (2)$$

Note that the updated message calculated in this way is unnormalized in general (in practice, the message is typically normalized explicitly after this equation is calculated).

Once the messages have converged after sufficient iterations of the message update equation, *beliefs*, i.e. estimates of the marginal probabilities of each variable, are calculated using the messages as follows:

$$b_i(x_i) = \frac{1}{Z_i} g(x_i) \prod_{k \in N(i)} m_{ki}(x_i) \qquad (3)$$

where $Z_i$ is a constant ensuring that the belief is normalized.

The main computational cost of BP arises from the message update (Eq. 2), specifically the difficulty of evaluating the sum in the right-hand side over $x_i$, which must be evaluated for each value of $x_j$ (i.e. the ``outer loop'' of the message update). If the variables $x_i$ can assume any of $M$ possible values (e.g. $M = 50$ possible integer disparities in a stereo problem), then updating the message from $i$ to $j$ requires $O(M^2)$ calculations, which can be very slow. But if the pairwise potential has the sparse compatibility property then we can greatly reduce this burden, as we describe in the next section.

## 4. The Trick

In general we can express the sparse compatibility condition as follows: if $x_i \notin Nbd(x_j)$, then $f(x_i, x_j) = \bar{f}$, where $Nbd(x_j)$ specifies the values of $x_i$ that are in a sparse neighborhood of $x_j$, i.e. the set of all $x_i$ that are compatible with $x_j$. In the stereo example above, $Nbd(x_j)$ is just the set of allowed disparities less than $T$ away from $x_i$. In other problems, the neighborhood structure may be more complicated; our technique applies even if the neighborhood function itself varies from node to node, i.e. $Nbd_i(x_j)$ rather than simply $Nbd(x_j)$.

Notice first that we can re-express the BP update equation as follows:

$$m_{ij}^{new}(x_j) = \sum_{x_i} f(x_i, x_j) h(x_i) \tag{4}$$

where

$$h(x_i) = g(x_i) \prod_{k \in N(i) \setminus j} m_{ki}(x_i) \tag{5}$$

Note that $h(x_i)$ is the product of all messages flowing into the $x_i$ node (except the message from $x_j$), multiplied by the unary potential $g(x_i)$ at that node.

We can rewrite the RHS of the equation by dividing the set of all $x_i$ (over which the sum is taken) into two pieces: $x_i \in Nbd(x_j)$ and $x_i \notin Nbd(x_j)$. Then the RHS becomes:

$$m_{ij}^{new}(x_j) = \sum_{x_i \in Nbd(x_j)} f(x_i, x_j) h(x_i) + \sum_{x_i \notin Nbd(x_j)} \bar{f} h(x_i) \tag{6}$$

since the pairwise potential has constant value $\bar{f}$ when $x_i \notin Nbd(x_j)$.

Then we can subtract and add the same quantity $A = \sum_{x_i \in Nbd(x_j)} \bar{f} h(x_i)$ as follows: subtract $A$ from the first term in Eq. 6 and add it to the second term. By adding $A$ to the second term, the sum in that term is extended over all $x_i$. The result is:

$$m_{ij}^{new}(x_j) = \sum_{x_i \in Nbd(x_j)} [f(x_i, x_j) - \bar{f}] h(x_i) + \bar{f} \sum_{x_i} h(x_i) \tag{7}$$

The two key points to notice are that the first term involves a sum over only a few states (the size of $Nbd(x_j)$), rather than over all $M$ states, and that the second term, $\bar{f} \sum_{x_i} h(x_i)$, is independent of $x_j$ and can be calculated once for all values of $x_j$ (i.e. it can be calculated in an outer loop). Thus, the only "inner loop" calculation is the first term, which requires only $O(mM)$ calculations, where $m$ is the size of $Nbd(x_j)$, and is much smaller than $M$.

Note also that the sparse compatibility property of the pairwise potential is slightly different from matrix sparseness in linear algebra: in our application $\bar{f}$ is generally non-zero, whereas the corresponding value in sparse matrices (the value of most matrix entries) is precisely zero. However, the trick can be considered to arise from the fact that if the pairwise potential is regarded as a matrix, it is the sum of a constant matrix (whose entries equal $\bar{f}$) and a sparse (in the linear algebra sense) matrix, and each of these component matrices give rise to fast message update calculations.

Finally, it is worth pointing out that our trick reduces to a very simple special case when $\bar{f} = 0$, i.e. when ``incompatible'' state pairs are completely disallowed rather than merely penalized (see the discussion of pruning in Sec. 2). In this case, the message update assumes the following very simple form:

$$m_{ij}^{new}(x_j) = \sum_{x_i \in Nbd(x_j)} f(x_i, x_j) h(x_i) \qquad (8)$$

which has the same complexity, $O(mM)$, as our trick. For some MRF models, setting $\bar{f}$ to $0$ may be a good approximation (or it naturally arise that $\bar{f} = 0$ exactly), but in many cases the non-negligible value of $\bar{f}$ is important (see Sec. 6 for an example).

## 5. Modification for Max-Product

With a small modification, the same idea also applies to the max-product form of BP (i.e. max-sum in the log domain). For max-sum, the message update equation is:

$$m_{ij}^{new}(x_j) = \max_{x_i}[f(x_i, x_j) + h(x_i)] \qquad (9)$$

where $h(x_i) = g(x_i) + \sum_{k \in N(i) \setminus j} m_{ki}(x_i)$.

The RHS can be expressed as

$$m_{ij}^{new}(x_j) = \max\{\max_{x_i \in Nbd(x_j)}[f(x_i, x_j) + h(x_i)], \max_{x_i \notin Nbd(x_j)}[\bar{f} + h(x_i)]\} \qquad (10)$$

Now, if we assume that $f(x_i, x_j) \geq \bar{f}$ for all pairs $(x_i, x_j)$ (this is usually the case for truncated potentials, although we didn't require this assumption for sum-product BP in the previous section), then we can expand the range over which the last $\max$ is calculated on the RHS of Eq. 10, thus obtaining the following result:

$$m_{ij}^{new}(x_j) = \max\{\max_{x_i \in Nbd(x_j)}[f(x_i, x_j) + h(x_i)], \max_{x_i}[\bar{f} + h(x_i)]\} \qquad (11)$$

Notice that the same speed-ups are obtained that arose in Eq. 7: the first term in the outermost $\max$ function can be efficiently calculated, and the second term is independent of $x_j$ and can be calculated once for all values of $x_j$ (i.e. it can be calculated in an outer loop).

## 6. Experimental Results

We demonstrate our approach on sum-product BP applied to a simple pairwise MRF stereo model, similar to that implemented by [2]. Our model is described briefly as follows. The unknown disparity at pixel $i$ is denoted by $x_i$, and the robust potential couples disparities between neighboring pixels $i$ and $j$: $f(x_i, x_j) = e^{-\alpha \min(|x_i - x_j|, T_b)}$, where $T_b$ is a maximum binary penalty strength. The unary potential measures how well a pixel $i$ in the right image matches the corresponding pixel in the left image. Given the left and right grayscale image intensities as a function of row and column, $L(r, c)$ and $R(r, c)$, and that the location of pixel $i$ is denoted by $(r_i, c_i)$, then a hypothetical disparity value, $x_i$, implies that we expect $R(r_i, c_i) \approx L(r_i, c_i + x_i)$. This is enforced by the following unary potential: $g(x_i) = e^{-\beta \min(|R(r_i,c_i)-L(r_i,c_i+x_i)|, T_u)}$, where $T_u$ is a maximum unary penalty strength.

The results of running ten sweeps (one sweep encompasses left, right, up and down message update directions across the entire pixel lattice) of BP, implemented in C++, on the standard Tsukuba image pair (resized to 144 x 192 pixels) [22] are shown in Fig. 2. The resulting disparity map was *identical* whether we ran standard BP or applied our speed-up technique, but our technique was approximately 3.1 times faster than standard BP (BP was run three times for each method: standard BP took a minimum of 21.09 sec./sweep and the sped-up BP took a minimum of 6.747 sec./sweep).

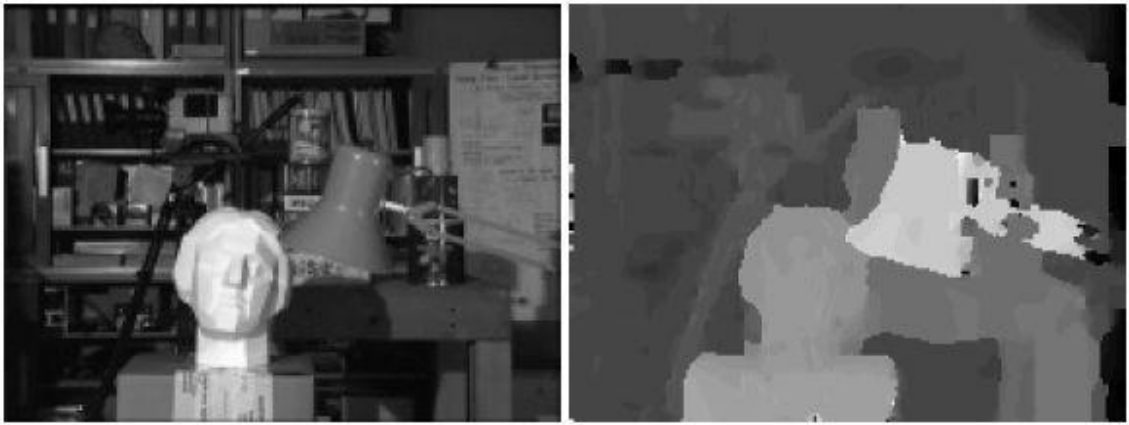

**Figure 2.** (a) Right image of Tsukuba stereo pair. (b) Disparity map estimated by BP is *identical* whether standard BP is used or our approach for speeding up BP.

We note that the resulting disparity map is significantly altered if we simplify the binary potential by setting $\bar{f} = 0$, i.e. preserving the functional form of $f(x_i, x_j)$ as above but setting $f(x_i, x_j) = 0$ for incompatible state pairs $(x_i, x_j)$. (This pruning approximation was discussed in Sec.'s 2 and 4.) The result of this approximation, shown in Fig. 3, demonstrates that for some models the non-negligible value of $\bar{f}$ may be important to the validity of the model.

Clearly, better stereo results could be obtained by making various improvements to the model (such as decreasing the binary potential penalty between pixels in high-gradient regions, e.g. [22]), but the point of the experiment is to validate the fact that our approach speeds up BP without altering the results. We note that speed-up factors of an order of magnitude or larger would result if the disparity range, relative to the size of $T_b$, were sufficiently large (such as would occur with wider baseline stereo image pairs). Further speed-ups to our sum-product BP algorithm could also be obtained by applying the multi-scale message passing or bipartite graph schemes described in [2] in tandem with our approach (keeping in mind that the distance transform tricks do not apply to sum-product BP).

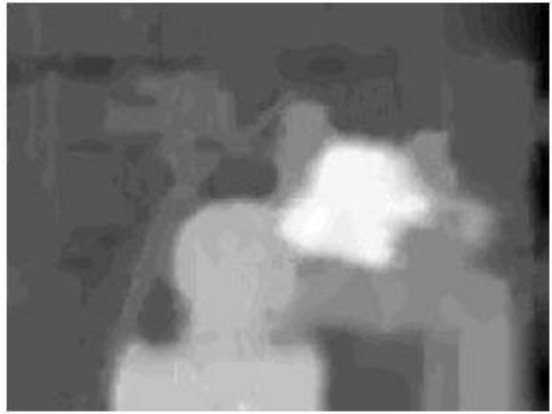

**Figure 3.** Result when $\bar{f}=0$, i.e. incompatible state pairs are disallowed (i.e. effectively pruned) rather than merely penalized by the pairwise potential. The quality of the disparity map is worse than that of the original model, demonstrating the possible pitfalls of approximating the pairwise potential in this way.

## 7. Conclusion

We have described a novel, exact and extremely simple technique for speeding up belief propagation for a variety of MRFs. The technique applies to any MRF with robust potential functions whose value is truncated to a fixed constant for sufficiently incompatible states (the sparse compatibility property). An important advantage of the method over most existing BP speed-ups is that it applies to sum-product BP, which is necessary for a variety of problems requiring marginal probability estimates (such as maximum likelihood approaches and CRFs), in addition to max-sum (max-product) BP. Experimental results on a stereo MRF model confirm that the approach confers a speed-up without altering the solution determined by BP.

While this paper focused on a certain common class of MRF models, we would like to point out that our technique applies to other types of graphical models for which other speed-up methods do not apply. One such category of models includes MRFs with robust pairwise potentials that are *not* functions of the difference of their two arguments, i.e. the functional form $f(x_i, x_j) = f(x_i - x_j)$ required by distance transform and FFT methods as in [2]. In addition, our technique applies to BP on factor graphs [11], in which the potential functions (i.e. factors) of two or more variables fulfill a sparse compatibility property (even if they don't fulfill the linearity requirement of [20]). For example, given a factor that is a function of three variables, $f(x_i, x_j, x_k)$, standard factor BP message updates will have complexity $O(M^3)$, where $M$ is the size of the variable state space; by contrast, our method will reduce the complexity to $O(C)$, where $C$ equals the size of the set of *compatible* state triplets $(x_i, x_j, x_k)$, which may be much smaller than the number of all possible state triplets, $M^3$.

Finally, since our technique exploits the fact that the pairwise potential matrix can be expressed as the sum of a constant matrix and a sparse matrix, we are exploring related methods of finding simple (e.g. low-rank) approximations of the pairwise potential matrix even when it doesn't fulfill the sparse compatibility property.

## Acknowledgments

We would like to acknowledge helpful feedback from Dr. Ender Tekin and Dr. Volodymyr Ivanchenko.